\documentclass[sigconf]{acmart}
\AtBeginDocument{%
  }

\setcopyright{acmlicensed}
\copyrightyear{2018}
\acmYear{2018}
\acmDOI{XXXXXXX.XXXXXXX}
\acmConference[Conference acronym 'XX]{Make sure to enter the correct
  conference title from your rights confirmation email}{June 03--05,
  2018}{Woodstock, NY}
\acmISBN{978-1-4503-XXXX-X/2018/06}

\usepackage{algorithm}
\usepackage{algpseudocode}
\usepackage{multirow}
\usepackage{booktabs}
\usepackage{bbding}
\usepackage{pifont}
\usepackage{utfsym}
\usepackage{amsmath}
\usepackage{makecell}
\usepackage{array}
\usepackage{amsmath}
\usepackage[inline]{enumitem}
\usepackage{graphicx}
\usepackage{adjustbox}
\usepackage{csquotes}
\usepackage{newfloat}
\usepackage{listings}
\usepackage{bm}
\usepackage{graphicx}
\usepackage{subcaption}
\usepackage{makecell}
\usepackage{threeparttable}
\usepackage{hyperref}

\usepackage{amsthm}

\usepackage{diagbox}

\usepackage{caption}
\usepackage{subcaption}

\begin{document}

\title{Generative Regression Based Watch Time Prediction for Short-Video Recommendation}

\author{Hongxu Ma}
\authornote{Both authors contributed equally to this research.}
\affiliation{%
  \institution{Fudan University}
  \city{Shanghai}
  \country{China}}
\email{hxma24@m.fudan.edu.cn}

\author{Kai Tian}
\authornotemark[1]
\affiliation{%
  \institution{KuaiShou Technology}
  \city{Beijing}
  \country{China}}
\email{tiank311@gmail.com}

\author{Tao Zhang}
\affiliation{%
  \institution{KuaiShou Technology}
  \city{Beijing}
  \country{China}}
\email{zhangtao08@kuaishou.com}

\author{Xuefeng Zhang}
\affiliation{%
  \institution{KuaiShou Technology}
  \city{Beijing}
  \country{China}}
\email{zhangxuefeng06@kuaishou.com}

\author{Han Zhou}
\affiliation{%
  \institution{Shanghai University of Finance and Economics}
  \city{Shanghai}
  \country{China}}
\email{zhouhan@stu.sufe.edu.cn}

\author{Chunjie	Chen}
\affiliation{%
  \institution{KuaiShou Technology}
  \city{Beijing}
  \country{China}}
\email{chencj517@gmail.com}

\author{Han Li}
\affiliation{%
  \institution{KuaiShou Technology}
  \city{Beijing}
  \country{China}}
\email{lihan08@kuaishou.com}

\author{Jihong Guan}
\affiliation{%
  \institution{Tongji University}
  \city{Shanghai}
  \country{China}}
\email{jhguan@tongji.edu.cn}

\author{Shuigeng Zhou}
\authornote{Corresponding author.}
\affiliation{%
  \institution{Fudan University}
  \city{Shanghai}
  \country{China}}
\email{sgzhou@fudan.edu.cn}

\begin{abstract}
Watch time prediction (WTP) has emerged as a pivotal task in short video recommendation systems, designed to quantify user engagement through continuous interaction modeling. 
Predicting users' watch times on videos often encounters fundamental challenges, including wide value ranges and imbalanced data distributions, which can lead to significant estimation bias when directly applying regression techniques.
Recent studies have attempted to address these issues by converting the continuous watch time estimation into an ordinal regression task.
While these methods demonstrate partial effectiveness, they exhibit notable limitations:
(1) the discretization process frequently relies on bucket partitioning, inherently reducing prediction flexibility and accuracy and (2) the interdependencies among different partition intervals remain underutilized, missing opportunities for effective error correction.

Inspired by language modeling paradigms, we propose a novel \textit{Generative Regression (GR)} framework that reformulates WTP as a sequence generation task. 
Our approach employs \textit{structural discretization} to enable nearly lossless value reconstruction while maintaining prediction fidelity. 
Through carefully designed vocabulary construction and label encoding schemes, each watch time is bijectively mapped to a token sequence. To mitigate the training-inference discrepancy caused by teacher-forcing, we introduce a \textit{curriculum learning with embedding mixup} strategy that gradually transitions from guided to free-generation modes. 

We evaluate our method against state-of-the-art approaches on two public datasets and one industrial dataset. 
We also perform online A/B testing on the Kuaishou App to confirm the real-world effectiveness. The results conclusively show that GR outperforms existing techniques significantly. 
Furthermore, we successfully apply GR to Lifetime Value (LTV) prediction, achieving 17.66\% MAE improvement over existing methods. These results validate GR as a generalizable solution for continuous value prediction tasks in recommendation systems.
\end{abstract}





\maketitle

\section{Introduction}

\begin{figure*}
    \centering
    \includegraphics[scale=0.65]{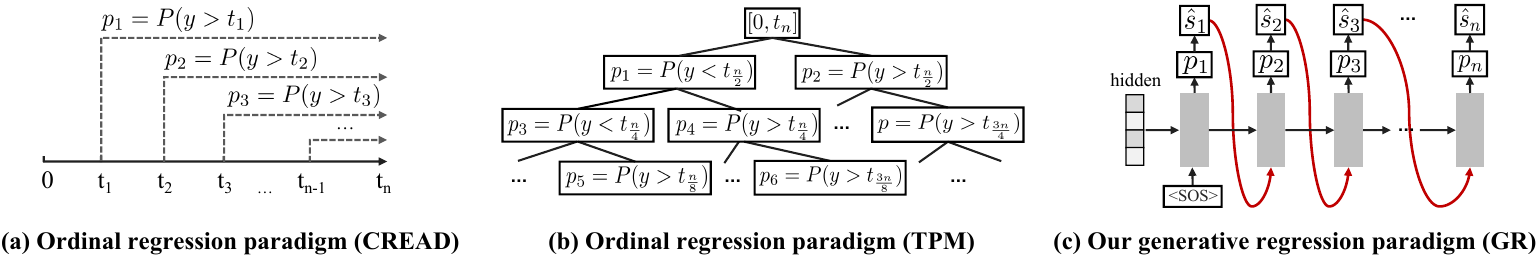}
    \caption{Predictive paradigm comparison among ordinal regression methods CREAD (a) and TPM (b), and our generative regression (c). Red lines indicate the discretization structure.}
    \label{fig:group}
    \vspace{-10pt}
\end{figure*}

        
       
        

In recent years, online short video content has a remarkable surge with the rapid development of short video social media platforms such as TikTok and Kuaishou, which spurs efforts to optimize recommendation systems for streaming players~\cite{covington2016deep, davidson2010youtube, liu2019user, liu2021concept}.
Unlike traditional Video on Demand (VOD) platforms such as Netflix and Hulu, short video platforms in scrolling mode automatically play content without the user clicking action for desirable video choice, rendering traditional metrics such as click-through rates obsolete~\cite{gao2022kuairand,gong2022real}.
Under these circumstances, the \textit{watch time} of videos has emerged as a critical metric for measuring user engagement and experience~\cite{covington2016deep, wang2020capturing, wu2018beyond, yi2014beyond}.
Continuous video watching means users' immersion and enjoyment of the platform, enhancing the probability of further user retention and conversion.
Consequently, accurate watch time estimation enables platforms to recommend videos prolonging users' viewing, which impacts key business metrics such as Daily Active Users (DAU) and drives revenue growth.
 
In contrast to limited and discrete actions such as liking, following, and sharing, \textit{watch time} generally exhibits a wide range and long-tailed distribution, making it fundamentally a regression problem for prediction.
Some methods~\cite{d2q,zhao2024counteracting,zheng2022dvr,zhao2023uncovering,zhang2023leveraging,tang2023counterfactual} optimize watch time prediction from a debiasing perspective but have not yet adequately addressed the core challenges of regression.
Some others~\cite{cread,tpm} transform the prediction problem into an Ordinal Regression (OR) task by employing a series of binary classifications across various predefined time intervals (buckets), as separately shown in Fig.~\ref{fig:group}(a) and Fig.~\ref{fig:group}(b). While effective, such a modeling paradigm still exhibits two major limitations as follows:

Firstly, conditional dependencies among time intervals are not fully leveraged, which are solely reflected in the definition of the labels. Predictions across different time intervals are often produced independently, thereby hindering the potential for effective error correction and leading to suboptimal results.
We provide rigorous theoretical proof of this limitation in the supplementary material.

Secondly, the strict discretization process within fixed time intervals in ordinal regression makes model performance highly contingent on the method of time interval segmentation, inherently reducing prediction flexibility and accuracy.
This approach performs binary classification across all predefined buckets, with the final prediction derived as the sum of bucket sigmoid probabilities multiplied by their corresponding bucket span values.
Due to the wide range of actual watch times~\cite{cread,tang2023counterfactual}, tail buckets often have excessively large span values, which can disproportionately amplify prediction errors for samples with shorter watch times, even when the binary probabilities of these tail buckets are minimal.
Additionally, the scrolling mode of short video platforms results in a high percentage of videos with relatively short watch times in real-world scenarios, further exacerbating the overall fitting error.

In response to these limitations above, inspired by the recent success of Large Language Models (LLMs)~\cite{touvron2023llama,brown2020language,zhao2023survey}, we propose a novel universal regression paradigm, called \textbf{G}enerative \textbf{R}egression (\textbf{GR}), which effectively utilizes dependencies among multi-step predictions and does not strictly rely on fixed time interval divisions. GR addresses the issues above as follows:

On the one hand, as shown in Fig.~\ref{fig:group}(c), 
the complete watch time prediction task is decomposed into a sequential generation task, where each step predicts a part of the total watch time.
The output of each time step serves as input for the next one, thereby constituting a conditional and sequential modeling process.
The objective is to predict a sequence of time slots, whose sum constitutes the continuous regression target. 
This generative regression paradigm not only ingeniously inherits the advantage of previous ordinal regression methods~\cite{cread, tpm, frank2001simple, li2006ordinal} by decomposing the regression task into multi-classification subtasks to simplify the prediction process, but also leverages dependencies between steps to accurately and progressively approximate the total watch time.

On the other hand, 
unlike ordinal regression methods~\cite{cread, tpm} that restrict outputs to binary classification within fixed time intervals,
our GR model offers the flexibility for each predictive step to not only select from a vocabulary of tokens—each representing a distinct time slot in positive real number space, but also output an end-of-sequence (<EOS>) token.
This flexibility enables GR to generate a broader set of potential sequences, thereby improving its capacity to generalize across diverse watching behaviors and leading to more accurate and personalized predictions. 

For token definition and watch time segmentation, we propose a data-driven unified vocabulary construction method, which mitigates token imbalance and eliminates manual design reliance, and a label encoding strategy allows a lossless restoration of watch time values, thereby enhancing the model's generality and generalization capability.
To accelerate model convergence, we adopt \textbf{curriculum learning}~\cite{bengio2009curriculum} strategy during training 
to alleviate training-and-inference inconsistency, commonly known as exposure bias~\cite{venkatraman2015improving, ding2017cold}.
Besides, leveraging our insights into the training process, we propose an \textbf{embedding mixup} method to compensate for output-to-input gradients. This approach enhances model performance at a lower computational cost by leveraging the semantic additivity of tokens while ensuring consistency between training and inference.
The contributions of this paper are as follows:
\begin{enumerate} [label=(\roman*)]
    \item We introduce a novel generation framework for predicting watch time, which inherits the benefits of structured discretization and adeptly utilizes interval relationships for the progressive and precise estimation of total watch time.
   \item To enhance generality and adaptability, we develop a data-driven unified vocabulary design and a label encoding method. Additionally, we introduce curriculum learning with embedding mixup to mitigate exposure bias and compensate for output-to-input gradients to accelerate model training.
    \item Extensive online and offline experiments show that GR significantly outperforms existing SOTA models. 
    We further analyze the underlying reasons for performance gain and the impact of key factors like vocabulary design to provide a clear understanding of the mechanisms underlying GR.
    \item Last but not least, we successfully apply GR to another regression task in recommendation systems, Lifetime Value (LTV) prediction, which indicates its potential as a novel and effective solution to general regression challenges.
\end{enumerate}

\section{Related Work}
\label{sec:related work}
\subsection{Watch Time Prediction (WTP)}
WTP aims to estimate the video watch time based on the user's profile, historical interactions, and video characteristics. 
Value regression (VR) directly predicts the absolute value of watch time, assessing model accuracy by mean square error~(MSE).
Subsequent WTP methods can be roughly divided into two groups. 
The first focuses on optimizing WTP from a debiasing perspective~\cite{d2q,zhao2024counteracting,zheng2022dvr,zhao2023uncovering,zhang2023leveraging,tang2023counterfactual}. CWM~\cite{zhao2024counteracting} introduces a counterfactual watch time, estimating a video's hypothetical full watch time to gauge user interest.
D2Co~\cite{zhao2023uncovering} differentiates actual user interest from duration bias and noisy watching using a duration-wise Gaussian mixture model.
However, these methods have not yet adequately addressed the core challenges of regression.
The second transforms the regression task into classification~\cite{cread, tpm, covington2016deep}.
CREAD~\cite{cread} introduces an error-adaptive discretization technique to construct dynamic time intervals. TPM~\cite{tpm} utilizes hierarchical labels to model relationships across varying granularities of time intervals.
Yet, these approaches are unable to fully capitalize on the interdependencies among these intervals and heavily rely on time interval segmentation.

\subsection{Ordinal Regression}
OR is a type of predictive modeling strategy employed when the outcome variable is ordinal
and the relative order of labels is important, such as age prediction~\cite{niu2016ordinal}, monocular depth perception~\cite{fu2018deep}, and head-pose estimation~\cite{hsu2018quatnet}. 
Recent works include specialized architectures like CNNOR~\cite{liu2017deep}, alternative training paradigms using soft labels such as SORD~\cite{diaz2019soft}, and dedicated probabilistic embedding methods~\cite{li2021learning}.
It has not been applied to watch time prediction until the introduction of CREAD~\cite{cread} and TPM~\cite{tpm}. These models decompose the regression task into multiple binary classification tasks, achieving significant benefits.

\begin{figure*}[t]
    \centering
    \includegraphics[width=0.99\linewidth]{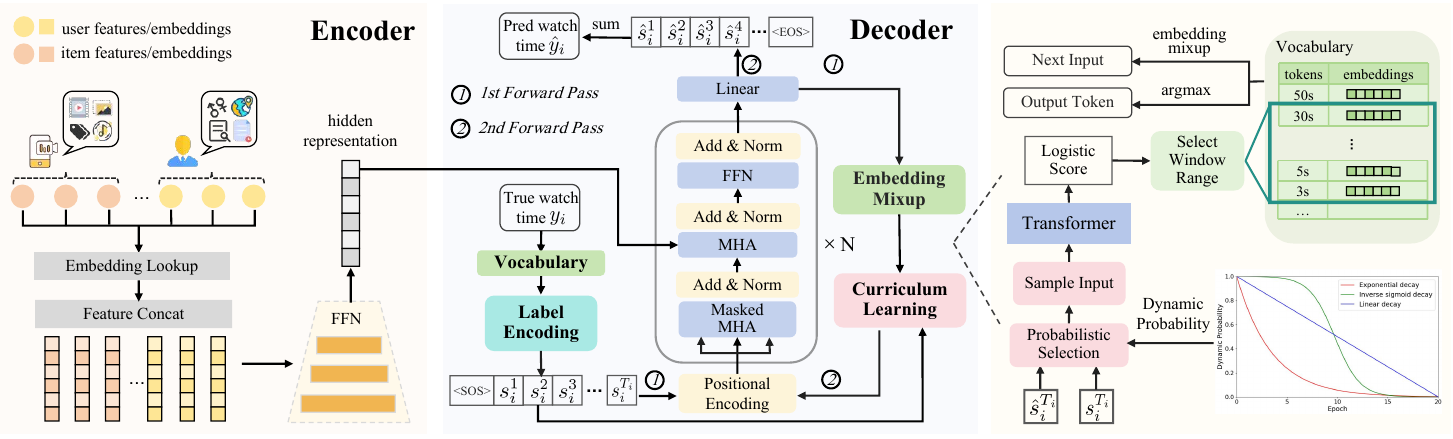}
    \caption{The framework of the GR model, which adopts an encoder-decoder architecture. The encoder extracts user and video features, while the decoder predicts watch time in an autoregressive manner and employs the curriculum learning with embedding mixup (CLEM) strategy to alleviate training-and-inference inconsistency introduced by teacher forcing.}
    \label{fig:framework}
\end{figure*}

\subsection{Sequence Generation}
Sequence generation learns contextual sequence mappings, initially prominent in NLP for tasks like machine translation~\cite{sutskever2014sequence,cho2014learning} and text summarization~\cite{bahdanau2014neural,vaswani2017attention}.
This paradigm extended to recommendation systems for capturing sequential user behavior patterns. 
In recommendation systems, sequential recommendation methods have been proposed to capture sequential patterns. 
GRU4Rec~\cite{hidasi2015session} is a session-based recommendation model with GRU.
SASRec~\cite{kang2018self} utilizes a self-attention mechanism to capture both long-term and short-term user preferences. 
BERT4Rec~\cite{sun2019bert4rec} employs a bidirectional transformer to encode item sequences. 
However, these sequential recommendation methods have predominantly focused on predicting the sequence of user behaviors, and their application to watch time prediction remains unexplored.

\section{Method}
\subsection{Problem Formulation}
Given a dataset $\mathcal{D}=\left\{ (\bm{u_i}, \bm{v_i}, y_i) \right\}_{i=1}^{N}$, where $\bm{u_i}$ and $\bm{v_i}$ represent the user-side features (such as user ID, static profile, and historical behaviors etc.) and the item-side (videos in this paper) features (e.g. tags, duration and category) of the $i$-th example respectively ~\footnote{We omit the context-side features for simplicity.}, $y_i\in \mathbb{R}$ is the corresponding watch time of the 
$i$-th example collected from recommendation system logs.
Value regression methods aim to learn a function  $f(\cdot)$ that directly maps the input features to a real-valued output, i.e., $y_i = f([\bm{u_i}; \bm{v_i}])$.

Sequence generation in \textbf{GR} is mostly based on autoregressive language modeling.
Specifically, we introduce a vocabulary $\mathcal{V} = \left\{ w_j \right\}_{j=1}^{V}$, where $V$ is the vocabulary size and each element $w_j$ represents a predefined time slot (e.g. 5 seconds, 10 seconds, etc.). The details of vocabulary construction are presented in Sec.~\ref{vocab_construction}.
Here, these time slots are analogous to \textit{tokens} in language models (LMs). Thus, ``token" and ``time slot" will be used interchangeably in the sequel.  
The vocabulary embedding matrix is denoted as $\bm{E} \in \mathbb{R}^{V \times D}$, where $D$ is the dimension of the time slot embeddings.

We decompose $y_i$ into a sequence of tokens $\bm{s_i} = \{s_i^1, ..., s_i^{T_i}\}$,
where $s_i^t\in \mathcal{V}$ and $T_i$ denotes the length of the sequence. This process, referred to as \textit{label encoding}, is described in detail in Sec.~\ref{label_encoding}. On the other hand, 
we design a \textit{label decoding} function $\text{g}(\cdot)$ \footnote{Here, $\text{g}(\cdot)$ functions as a lookup table that maps tokens to real-valued vocabulary entries, e.g., $\text{g}(``\text{30s}") = 30$.} that reconstructs the original watch time $y_i$ from $\bm{s_i}$, i.e., $y_i=\text{g}(\bm{s_i})=\sum_{t=1}^{T_i}\text{g}(s_i^t)\in \mathbb{R}$.  
Our goal is to train a sequence generation model, given user and video characteristics $(\bm{u_i}, \bm{v_i})$, which generates the corresponding sequence of watch time slots $\bm{\hat{s}_i} = \{ \hat{s}_i^1, \hat{s}_i^2, ..., \hat{s}_i^{T_i}\}$, and in turn,  from which the predicted watch time $\hat{y_i}=\text{g}(\bm{\hat{s}_i})=\sum_{t=1}^{T_i} \text{g}(\hat{s}_i^t)$ approximates the actual watch time $y_i$.

\subsection{The Generative Regression (GR) Model}
As shown in Fig.~\ref{fig:framework}, \textbf{GR} adopts a Transformer-based encoder-decoder architecture. 
The encoder extracts user and video features, while the decoder predicts the watch time in an autoregressive manner. 

\subsubsection{Encoder} 
Unlike traditional sequence-to-sequence tasks or user behavior modeling in recommendation systems, watch time prediction does not inherently depend on the order of user history interacted items. 
To ensure model generality and simplicity, we follow previous works~\cite{tpm, cread} and employ a feedforward network (FFN) as an encoder. Note that this encoder can be replaced with \textit{any sophisticated} model architecture.
Formally, the encoder extracts user and video features to produce a fixed-length hidden feature $\bm{h_i} \in \mathbb{R}^{1 \times D}$ that will be fed to the decoder as follows:
\begin{equation}
    \bm{h_i} = \bm{W_L}\cdot(... \text{relu}(\bm{W_2}\cdot(\text{relu}(\bm{W_1}\cdot\bm{x}_i))))
\end{equation}
where $\bm{x_i}=[\bm{u_i};\bm{v_i}]$, $\bm{W_1},...,\bm{W_L}$ are weight parameters of FFN.

\subsubsection{Decoder}
The decoder adopts a Transformer~\cite{vaswani2017attention} architecture, comprising standard Transformer blocks.
Each block contains Masked Multi-Head Self-Attention (Masked MHA), Multi-Head Cross-Attention (MHA), and a position-wise Feed-Forward Network (FFN).
To reduce computational overhead, we employ a simplified hyperparameter configuration, with detailed hyperparameter settings provided in the supplementary material.
As in language modeling, we introduce three special tokens into the vocabulary $\mathcal{V}$:  $\text{<SOS>}$, $\text{<EOS>}$ and $\text{<PAD>}$ represent start-of-sequence token, end-of-sequence token, and padding token, respectively.
For each target sequence $\bm{s_i}$,  $\text{<SOS>}$ and $\text{<EOS>}$ will be added to the start and the end of the sequence. 
The $\text{<PAD>}$ token is used to pad sequences within a batch to have the same length, facilitating efficient parallel computation.
As these tokens do not represent any meaning in the label space (i.e., $\text{g}(c)=0, c\in \{\text{<SOS>}, \text{<EOS>}, \text{<PAD>}\}$), we will omit these tokens in our math formulation for better understanding.

As illustrated in~Fig.\ref{fig:framework}, the decoder generates the sequence of watch time slots $\bm{\hat{s_i}} = \{ \hat{s}_i^1, ..., \hat{s}_i^j, ... , \hat{s}_i^{T_i}\}$ conditioned on the encoder output $\bm{h_i}$ and the preceding subsequence. 
Specifically, at time step $t$ in training, the output token $\hat{s}_i^t$ will be computed as

\begin{equation}
    \hat{s}_i^t = \arg\max_{w\in \mathcal{V}} P_{\theta}(w \mid \bm{h_i}, \bm{\hat{s}_i^{<t}})
    \label{eq:hat_w}
\end{equation}
where $\theta$ is the model parameter and $\bm{\hat{s}_i^{<t}}$  represents the tokens generated before.
Utilizing the chain rule, the overall probability of generating the sequence can be expressed as
\begin{equation}
    P_{\theta}(\bm{s_i} \mid \bm{h_i}) = P_{\theta}( s_i^1, ..., s_i^{T_i} \mid \bm{h_i})= \prod_{t=1}^{T} P_{\theta}(s_i^t \mid \bm{h_i}, \bm{{s}_i^{<t}})
    \label{eq:chain_rule}
\end{equation}

Three key issues remain to be addressed: (1) how to construct an effective vocabulary, (2) how to encode $y_i$ into a sequence $\bm{s_i}$, and (3) how to optimize the model. These issues are detailed in the following sections. 

\subsection{Vocabulary Construction}
\label{vocab_construction}



As mentioned before, tokens in vocabulary $\mathcal{V}$ represent predefined watch time slots that enable the model to generate sequences closely approximating the actual watch time values.
Based on our cognition of the deep regression task, three principles are designed to guide the construction of vocabulary.

\begin{itemize}[itemindent=0pt, left=4pt]
    \item \textbf{Completeness}: The vocabulary $\mathcal{V}$ must be able to represent all watch time values $\{y_i\}_{i=1}^N$ using a finite number of tokens almost without loss.
    Also, each token must be unique.
     \item \textbf{Balance}: The frequencies of tokens should be relatively uniform to prevent class imbalance.
    \item \textbf{Adaptability}: The vocabulary should remain consistent to ensure scalability and adaptability across various datasets.
\end{itemize}

One intuitive strategy is to select watch time values from the dataset as tokens based on several fixed percentiles, yet failing to meet the completeness principle. 
An alternative is to select watch time values as tokens based on one fixed percentile, then subtract the token values from all watch time values that exceed them, repeating this process until the residuals become negligible, which fails to meet the balance principle. Due to the space limit, details of this strategy are provided in the supplementary materials.

\begin{algorithm}[t]
\caption{Constructing Vocabulary with dynamic percentiles}
\small
\label{algorithm:dynamic}
\begin{algorithmic}[1]
\Require Dataset labels $\bm{Y} = \{y_j\}_{j=1}^N$, initially empty Vocabulary $\mathcal{V} = \{\}$, 
start percentile $q_{\text{start}}$, end percentile $q_{\text{end}}$, decay rate $\alpha$, minimal restoration error $\epsilon$.
\State Sort $\bm{Y}$ in descending order to obtain $\hat{\bm{Y}} = \{\hat{y}_j\}_{j=1}^N$.
\State Initialize iteration counter $i = 1$, error metric $err = \infty$, current percentile $q = q_{\text{start}}$
\While{$err > \epsilon$}
    \State Compute the $q$-percentile $o_i$ of $\hat{\bm{Y}}$.
    \If{$o_i = 0$} \Comment{Terminate if the percentile value is zero}
        \State break
    \EndIf
    \State Generate a new token $v_i$ which satisfy $o_i=g(v_i)$ and insert $v_i$ into vocabulary $\mathcal{V}$.
    \State Update $\hat{\bm{Y}}$ using:
    \[
    \hat{y}_j = 
    \begin{cases}
        \hat{y}_j, & \text{if } \hat{y}_j < o_i, \\
        \hat{y}_j - o_i, & \text{otherwise}
    \end{cases}
    \]
    \State Update the error metric $err$: $err = \max\{\frac{\hat{y}_j}{y_j}\}_{j=1}^N $
    \State Update percentile $q$ with decay rate $\alpha$: $q = \max(q \cdot \alpha,~ q_{\text{end}})$
    \State Increase $i$: $i=i+1$.
\EndWhile
\State \Return $\mathcal{V}$
\end{algorithmic}
\end{algorithm}

To address both principles simultaneously, we propose a data-driven vocabulary construction algorithm using \textit{dynamic quantile adjustment} (Algorithm.~\ref{algorithm:dynamic}).
The algorithm initializes with a high starting quantile $q_{start}$ and adaptively reduces it by decay rate $\alpha$ until reaching the terminal quantile $q_{end}$. 
This strategy expedites the reduction of tail values, rapidly decreasing the variance among updated values, which effectively mitigates the challenges posed by the long-tailed distribution in the dataset, for which we provide detailed experimental validation in Sec.~\ref{sec:RQ3}.

We emphasize that our vocabulary construction and label encoding process, while analogous to linguistic syntax building for sequence generation, does not presume theoretical optimality. The proposed strategy serves as a principled engineering solution, leaving theoretical analysis of optimal tokenization for future work.

\subsection{Label Encoding}
\label{label_encoding}
Given the vocabulary $\mathcal{V}=\{w_1, w_2, ..., w_V\}$, we perform label encoding to transform the watch time values $\{y_i\}_{i=1}^N$ into corresponding target sequences $\{\bm{s}_i = \{s_i^1, \ldots, s_i^{T_i}\}\}_{i=1}^N$.
To guide the label encoding process, we propose three foundational principles:
\begin{itemize}[itemindent=0pt, left=4pt]
    \item \textbf{Correctness}: The original value must be reconstructible from the token sequence with bounded error:
    \begin{equation}
        y_i = \sum_{t=1}^{T_i} g(s_i^t) + \epsilon,\quad \text{where } |\epsilon| \leq 0.001 \cdot y_i
    \end{equation}
    \item \textbf{Minimal Sequence Length}: The sequence length $T_i$ should achieve the minimal possible cardinality while satisfying the correctness constraint.
    \item \textbf{Monotonicity}: Token values must satisfy a non-increasing order:
    \begin{equation}
        g(s_i^1) \geq g(s_i^2) \geq \cdots \geq g(s_i^{T_i})
    \end{equation}
\end{itemize}
The minimum sequence length principle reduces learning complexity, while the monotonic constraint captures decaying user attention patterns during video watching.

To follow these principles, we implement a greedy decomposition algorithm. Starting from the largest possible watch time slot and decreasing progressively, decomposing the total watch time $y_i$ into a sequence of watch time slots.

\subsection{Optimization and Inference}
\subsubsection{Vanilla Training Process}
Following language modeling paradigms, the model predicts the next token $s^t$ conditioned on preceding ground truth tokens $s^{<t}$. The learning objective minimizes the cross-entropy loss between predicted and ground truth sequences:
\begin{equation}
    \mathcal{L}_{ce} = -\sum_{i=1}^{N} \sum_{t=1}^{T_i} \log P_{\theta}(\hat{s}_i^t \mid \bm{h_i}, \hat{s}_i^{<t})
    \label{eq:loss_ce}
\end{equation}
 
Following previous works~\cite{tpm, cread}, we employ the Huber loss~\cite{huber1992robust} to guide regression: 
\begin{equation}
    \mathcal{L}_{huber} = \mathcal{L}_{\delta}(y_i, \hat{y}_i) = \begin{cases} 
\frac{1}{2}(y_i - \hat{y}_i)^2 & \text{if} |y_i - \hat{y}_i| \leq \delta, \\
\delta \cdot (|y_i - \hat{y}_i| - \frac{1}{2} \delta) & \text{otherwise}
\end{cases}
\end{equation}
where $\hat{y_i} = \sum_{t=1}^{T_i} g(\hat{s}_i^t)$, $\delta$ acts as a threshold, toggling between quadratic and linear losses to balance sensitivity and robustness against outliers.
Therefore, the composite loss becomes:
\begin{equation}
\mathcal{L}=\mathcal{L}_{ce}+\lambda \cdot \mathcal{L}_{huber}
\label{eq:total_loss}
\end{equation}
where $\lambda$ is a hyperparameter that balances the two losses.
To improve model efficiency, we adopt a teacher forcing (TF) strategy~\cite{venkatraman2015improving}, which directly feeds the ground truth output $s_{i}^{t}$ as input at step $t+1$ to guide model training.
However, since the ground truth is unknown during inference, the discrepancy of input for the decoder leads to the well-known exposure bias problem~\cite{goodman2020teaforn}, which can degrade model performance. 

\subsubsection{Curriculum Learning with Embedding Mixup (CLEM)}
To mitigate exposure bias inherent in teacher forcing, we propose a phased \textbf{Curriculum Learning (CL)} strategy. 
Specifically, to predict $\hat{s}_i^t$, we randomly choose ground truth tokens $s_i^{t-1}$ or predicted tokens $\hat{s}_i^{t-1}$ with a dynamic probability $p$ as the sampling rate.
However, Transformer processes the entire sequence in parallel during a single forward pass, preventing access to the predicted tokens of previous time steps.
Thus, as shown in Fig.~\ref{fig:framework}, we implement CL with \textbf{two forward passes} through the decoder during training.
The first pass performs vanilla training to obtain initial model predictions. 
In the second pass, inputs are sampled between ground truth tokens and predicted tokens with probability $p$, yielding the final predictions. Both passes share the same model parameters.



To warm up, we start with $p\approx 1$, indicating that the model predominantly relies on the ground truth tokens. 
We then adjust the probability $p$ using a non-linear decay strategy, which increases the likelihood of sampling from the predicted sequence. 
This enables the model to gradually adapt to the inference stage. Formally, 
\begin{equation}
        p = p_{0} \cdot \frac{\omega}{\omega + e^{\left(\frac{\tau}{\omega}\right)}}
        \label{eq:cl}
    \end{equation}
where $\tau$ is the training iteration and $\omega>0$ influences the shape of the descent curve to ensure a seamless transition from higher to lower values.
This strategy addresses exposure bias by learning to predict with both ground truth and previous prediction as input.
In Sec.~\ref{sec:ablation}, we also conduct detailed experimental comparisons of additional strategies such as linear and exponential decay.

\begin{figure}[t]
    \centering
    \begin{subfigure}[b]{0.49\linewidth}
        \includegraphics[width=\linewidth]{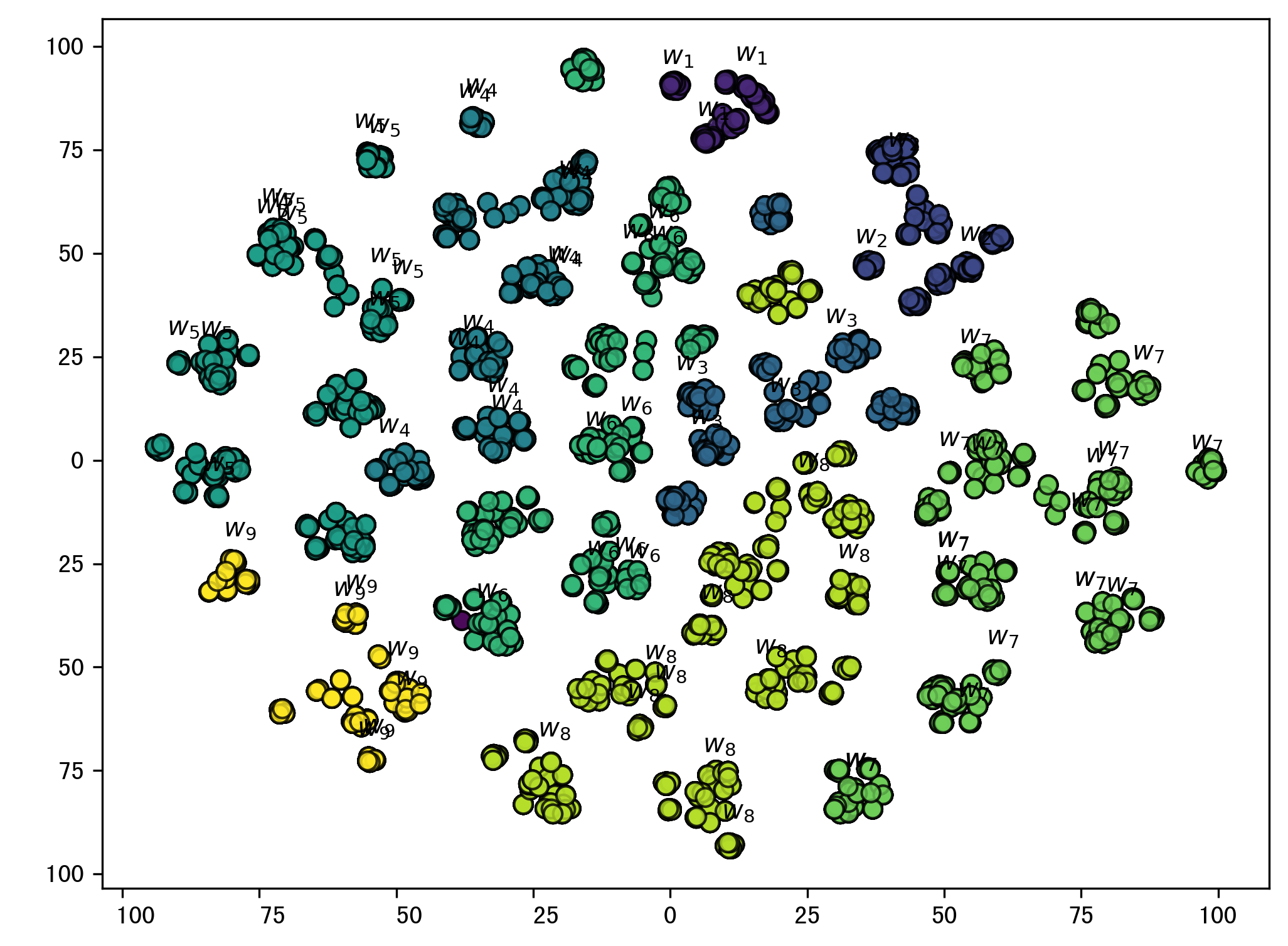}
        \label{fig:num_emb_vis}
       \vspace{-15pt}
        \caption{Watch time embeddings visualization during training.}
    \end{subfigure}
    \begin{subfigure}[b]{0.45\linewidth}
        \includegraphics[width=\linewidth]{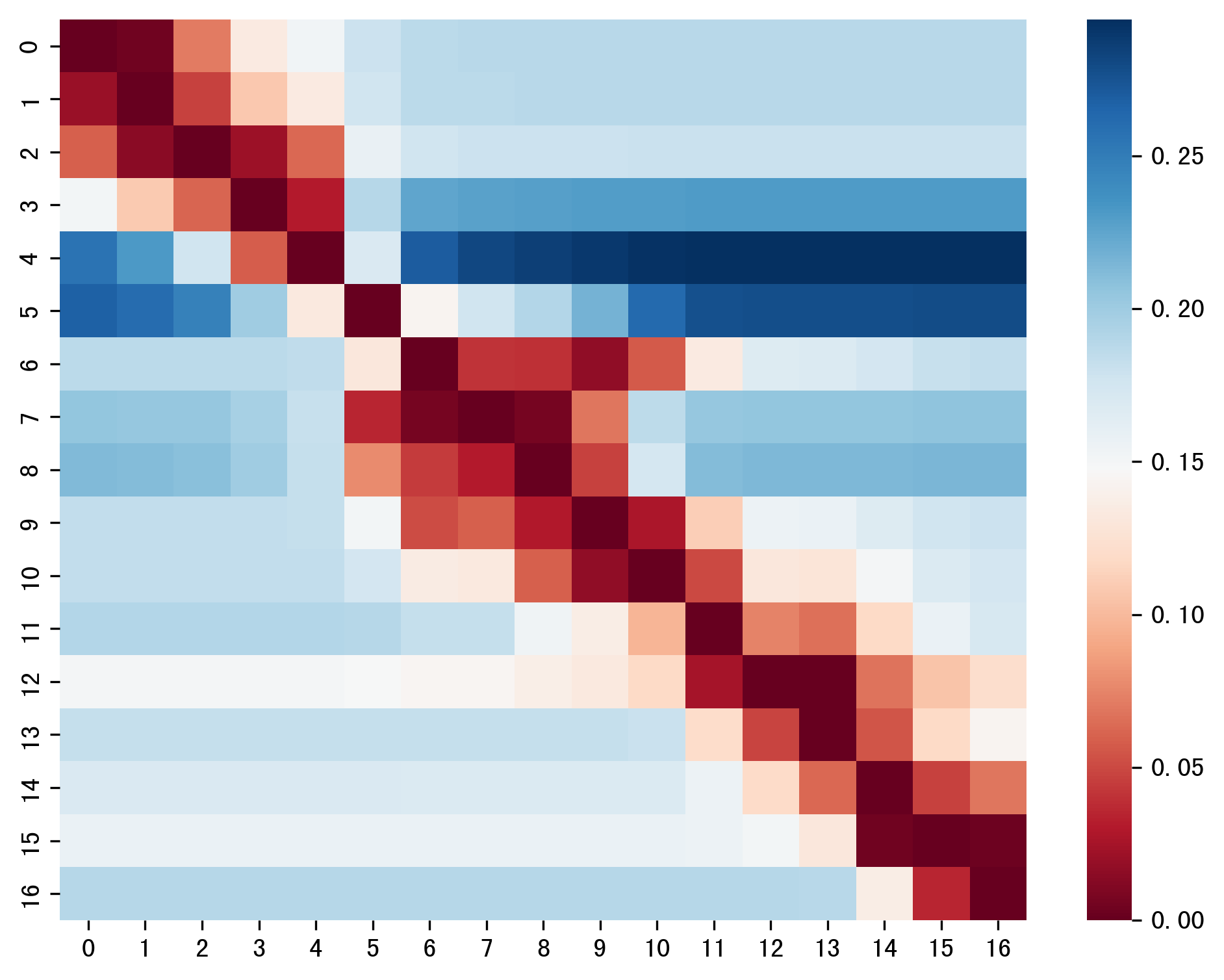}
        \label{fig:prob_diff}
       \vspace{-15pt}
        \caption{Probability difference score among tokens during training.}
    \end{subfigure}
    \caption{Watch time embedding with a weighted sum of token embeddings (left) and the probability distribution difference among tokens for each $\hat{s}_i^t$ (right). Best viewed in color.}
    \label{fig:vis}
\end{figure}

Our analysis reveals that GR effectively captures inter-token relationships through its embedding structure. Given the vocabulary size $V$ being orders of magnitude smaller than typical language models, we analyze token semantics via aggregated value embeddings:
\begin{equation}
    \bm{e}_i = \sum_{t=1}^{T_i} r_t \bm{E}[s_i^t,:], \quad r_t = \frac{g(s_i^t)}{y_i}
\end{equation}

where $r_t$ represents the contribution of token $s_i^t$ to target value $y_i$. Fig.~\ref{fig:vis}(a) demonstrates two key properties:

\begin{itemize}[left=4pt]
    \item \textbf{Token Clustering}: Values sharing initial tokens form distinct clusters.
    \item \textbf{Semantic Continuity}: Embeddings of tokens with similar $g(w_j)$ values reside in proximate regions.
\end{itemize}

This structural coherence facilitates numerical reasoning through geometrically meaningful representations.
As noted in Sec.~\ref{label_encoding}, 
tokens are arranged in non-increasing order within the vocabulary $\text{g}(w_1)> \text{g}(w_2)> ... > \text{g}(w_V)$. 
We also compute the averaged probability difference of each token relative to its neighbors and observe that tokens with neighboring indices in the vocabulary demonstrate the highest probability similarity in the model's predictions, as shown in Fig.~\ref{fig:vis}(b). 



To improve the prediction precision of next token, we propose to integrate the embedding sequences of the preceding tokens through a local ensemble approach called \textbf{Embedding Mixup (EM)} during the training process.
The cohort is centered on the current predicted token $\hat{s}^{t}_i$ with window size $n_w$, the mixup region is $[\delta_{\hat{s}^{t}_i}-b, \delta_{\hat{s}^{t}_i}+b]$ and $b=\lfloor\frac{n_w}{2}\rfloor$, $\delta_{\hat{s}^{t}_i}$ represents the token index of $\hat{s}^{t}_i$ in $\mathcal{V}$. We have
\begin{equation}
    \bm{z_i}^{t}=\sum\limits_{j=0}^{n_w}\sigma_{j}\cdot \bm{E}[\delta_{\hat{s}^{t}_i}+j-b,:]
\label{eq:cf_main}
\end{equation}
\begin{equation}
    \sigma_{j}=\frac{exp(-\rho_{j})}{\sum_{k=0}^{n_w}exp(\rho_{\delta_{\hat{s}^{t}_i}+k-b})}
\end{equation}
where $\bm{E} \in \mathbb{R}^{V \times D}$ is the vocabulary embedding matrix, $\sigma_{j}$ recalculates the fusion weights of tokens in the fixed window size, $\rho_j$ is the logit predicted by the decoder at step $t$.
Specifically, $\bm{z_i}^{t}$ will replace original $\bm{E}[\hat{s}^{t}_i,:]$ as the input at $t+1$ step.
This re-weighted EM approach leverages the numerical semantics and additivity property inherent in our tokens. The re-weighting ensures that the semantic space is aligned, eliminating any discrepancies in scale.
EM offers three benefits: 1) It reduces the learning complexity of the model by merging representations of tokens within a fixed window, thereby preventing significant errors; 2) The integration leverages the predicted scores from the previous steps, enhancing the information transfer from output to input in recurrence structure and restructuring the gradient propagation path; and 3) it ensures consistency between training and inference while lowering inference cost.

\subsubsection{Inference Process}
During inference, the encoder extracts $\bm{h_i}$ from input features $[\bm{u_i}, \bm{v_i}]$, the decoder begins with the $\text{<SOS>}$ token and sequentially generates the prediction sequence $\hat{s}_i = \{ \hat{s}_i^1, \hat{s}_i^2, ..., \hat{s}_i^{T_i}\}$,  with each token generated using only \textbf{the first forward pass}.
The process continues until the token $\text{<EOS>}$ is generated, which signifies the completion of the sequence.
Finally, the predicted watch time is computed as $\hat{y_i} = \sum_{t=1}^{T_i} \text{g}(\hat{s}_i^t)$.

\section{Experiments}
This section presents extensive experiments to demonstrate the effectiveness of the GR model.
Five research questions are explored in these experiments:

\begin{itemize}[itemindent=0pt, left=4pt]
    \item \textbf{RQ1:} How does GR compare to state-of-the-art methods in terms of prediction accuracy of watch time? 
    \item \textbf{RQ2:} What are the underlying reasons behind the model's performance exceeding the baseline?
    \item \textbf{RQ3:} What is the effect of vocabulary design on the performance of GR and why? 
    \item \textbf{RQ4:} What impact does CLEM have on the GR model, and how do different training strategies affect performance? 
    \item \textbf{RQ5:} How does GR perform on other regression tasks?
\end{itemize}


\subsection{Experiment Settings}
\subsubsection{Datasets.}
We evaluate our method on one industrial dataset and two public benchmarks.
The large-scale industrial dataset (\textbf{Indust} for short) is sourced from a real-world short-video app Kuaishou with over 400 million DAUs and multi-billion impressions each day.
We use interaction logs spanning 4 days for training and those from the subsequent day for testing.
We also use the public \textbf{CIKM16}\footnote{https://competitions.codalab.org/competitions/11161} and \textbf{KuaiRec}~\cite{gao2022kuairec} datasets, adopting the experimental settings from previous works~\cite{tpm, cread} (Details are provided in the supplementary material).
Consistent with prior work~\cite{tpm}, we also report watch ratio results on KuaiRec, which can be used in conjunction with video duration to calculate watch time. 

\subsubsection{Metrics}
To evaluate the proposed method, we follow previous work~\cite{tpm,cread} and utilize two performance metrics:

\begin{itemize}[itemindent=0pt, left=4pt]
    \item \textbf{Mean Average Error (MAE)}: It quantifies regression accuracy by averaging the absolute deviations between predicted values $\{\hat{y_i}\}_{i=1}^N$ and actual values $\{y_i\}_{i=1}^N$ by $\frac{1}{N} \sum_{i=1}^{N} \left| \hat{y}_i - y_i \right|$.

    \item \textbf{XAUC~\cite{d2q}}: This measure assesses the concordance between the predicted and actual ordering of watch time values. We uniformly sample pairs from the test set and calculate the XAUC by determining the percentile of samples that are correctly ordered. A higher XAUC indicates better model performance.
\end{itemize}

\begin{table*}[ht]
\centering
\caption{Performance comparison among different approaches 
on KuaiRec, CIKM16 and Indust dataset.}
\label{tab:watch_time}
\resizebox{\textwidth}{!}{
\begin{tabular}{c|ccc|ccc|ccc|cc}
\hline
\multirow{2}{*}{Method} & \multicolumn{3}{c|}{KuaiRec~(watch time)} & \multicolumn{3}{c|}{KuaiRec~(watch ratio)} & \multicolumn{3}{c|}{CIKM16} & \multicolumn{2}{c}{Indust} \\
& MAE~$\downarrow$ & XAUC~$\uparrow$ & XAUC Improv. & MAE~$\downarrow$ & XAUC~$\uparrow$ & XAUC Improv. & MAE~$\downarrow$ & XAUC~$\uparrow$ & XAUC Improv. & MAE~$\downarrow$ & XAUC~$\uparrow$ \\
\hline
VR & 7.634 & 0.534 & - & 0.385 & 0.691 & - & 1.039 & 0.641 & - & 46.343 & 0.588 \\
WLR~\cite{covington2016deep} & 6.047 & 0.545 & 2.059\% & 0.375 & 0.698 & 1.013\% & 0.998 & 0.672 & 4.836\% & - & -\\
D2Q~\cite{d2q} & 5.426 & 0.565 & 8.757\% & 0.371 & 0.712 & 3.039\% & 0.899 & 0.661 & 3.120\% & - & -\\
CWM~\cite{zhao2024counteracting} & 3.452 & 0.580 & 8.614\% & 0.368 & 0.725 & 4.920\% & 0.891  & 0.662  & 3.276 \% & - & - \\
TPM~\cite{tpm} & 3.456 & 0.571 & 6.929\% & \underline{0.361} & 0.734 & 6.223\% & \underline{0.850} & 0.676 & 5.460\% & 41.486 & 0.593\\
CREAD~\cite{cread} & \underline{3.307} & \underline{0.594} & \underline{11.236\%} & 0.369 & \underline{0.738} & \underline{6.802\%} & 0.865 & \underline{0.678} & \underline{5.772\%} & \underline{39.979} & \underline{0.597} \\
\textbf{GR (ours)} & \textbf{3.196} & \textbf{0.614} & \textbf{14.981\%} & \textbf{0.333} & \textbf{0.753} & \textbf{8.972\%} & \textbf{0.815} & \textbf{0.691} & \textbf{7.80\%} & \textbf{38.528} & \textbf{0.604} \\
\hline
\end{tabular}
}
\begin{tablenotes}\footnotesize
\item[*] Here, the best and second best results are marked in \textbf{bold} and \underline{underline}, respectively. 
$\uparrow$ indicates that the higher the value is, the better the performance is, while $\downarrow$ signifies the opposite.
Each experiment is repeated 5 times and the average is reported.
\end{tablenotes}
\end{table*}

\begin{table}[t]
    \centering
    \small
    \caption{Performance gain on online A/B testing.}
    \label{tab:ab_test}
    \vspace{-1em}
    \resizebox{0.44\textwidth}{!}{
        \begin{tabular}{c|c|c}
            \hline
            \multirow{3}{*}{A/B test} & APP Usage Time & +0.112\% (p-value=0.01) \\
            \cline{2-3}
            & Average App Usage Per User & +0.087\% \\
            \cline{2-3}
            & Video Consumption Time & +0.129\% \\
            \hline
        \end{tabular}
    }
\begin{tablenotes}\footnotesize
\item[*] In a stable video recommendation system, a \textbf{0.1\% increase} is significant.
\end{tablenotes}
\end{table}


\begin{table}[t]
    \centering
    \caption{Comparison of vocabulary construction methods.}
    \label{tab:vocabulary}
    \vspace{-1em}
    \resizebox{0.4\textwidth}{!}{
        \begin{tabular}{c|cc|cc}
            \hline
            \multirow{2}{*}{Vocabulary design} & \multicolumn{2}{c|}{KuaiRec} & \multicolumn{2}{c}{CIKM16} \\
            & MAE~$\downarrow$ & XAUC~$\uparrow$ & MAE~$\downarrow$ & XAUC~$\uparrow$ \\
            \hline
            Manual & 3.281 & 0.604 & 0.825 & 0.685  \\
            Binary & 3.268 & 0.605 & 0.821 & 0.687  \\
            \textbf{Dynamic quantile} & \textbf{3.196} & \textbf{0.614} & \textbf{0.815} & \textbf{0.691} \\
            \hline
        \end{tabular}
    }
\end{table}


\subsubsection{Compared Methods}
Considering baseline methods compared in prior studies~\cite{cread, tpm}, we compare several state-of-the-art methods~\cite{covington2016deep,d2q, zhao2024counteracting, cread, tpm} with our GR. 
More details of the compared methods are provided in the supplementary material.

\subsection{Performance Comparison (RQ1)}
\label{sec:RQ1}
\subsubsection{Offline Evaluation}
Tab.~\ref{tab:watch_time} shows the comparative results between GR and six baselines across three datasets. GR achieves consistent improvements in both MAE and XAUC metrics. For watch time prediction, GR maintains superior performance with {4.117\%} MAE reduction and a {1.917\%} XAUC improvement on CIKM16. 
On the KuaiRec, it significantly outperforms the second-best method with a {3.356\%} MAE reduction and {3.367\%} XAUC lift.
As for Indust dataset, GR exhibits a {3.629\%} relative decrease in MAE and a {1.001\%} improvement in XAUC compared to the CREAD, which is a notable enhancement on a real-world business dataset.
Regarding watch ratio predictions, 
while all models gain significantly from eliminating duration bias,
GR maintains the best performance, boasting a {7.756\%} MAE reduction and a {2.033\%} XAUC improvement.
The comprehensive improvements in both MAE and XAUC substantiate GR's superiority. 
We also conduct experiments with parameter-equivalent models (see supplementary materials) to ensure the performance gains are not solely from increased model parameters.

\subsubsection{Online A/B Testing}
We also conduct an online A/B test on the Kuaishou App to demonstrate the real-world efficacy of our method. Considering that Kuaishou serves over 400 million users daily, doing experiments from 6\% of traffic involves a huge population of more than 25 million users, which can yield highly reliable results. The predicted watch times are used in the ranking stage to prioritize items with higher predicted watch times, making them more likely to be recommended.
The online experiment has been launched on the system for six days, with evaluation metrics including app usage time, average app usage per user, daily active users, and video consumption time (accumulated watch time). The control group utilized the CREAD model, while the proposed GR framework exhibited a 10.2\% reduction in average queries per second (QPS) during online serving. Despite this computational overhead, the overall return on investment (ROI) met the threshold for full deployment, indicating favorable trade-offs between operational costs and business value enhancement.
As shown in Tab.~\ref{tab:ab_test}, the results demonstrate that GR consistently boosts performance in watch time related metrics, with an improvement by 0.087\% on average app usage per user, significant \textbf{0.129\%} on video consumption time and \textbf{0.112\%} on app usage time with $\text{p-value}\footnote{Lower p-values mean greater statistical significance (e.g., p=0.01 implies a 1\% likelihood of gain occurring by chance).} = 0.01$, substantiating its potential to significantly enhance real-world user experiences.


\subsection{Underlying Reasons Analysis For Performance Gain (RQ2)}
\begin{figure}[t]
    \centering
    \includegraphics[scale=0.38]{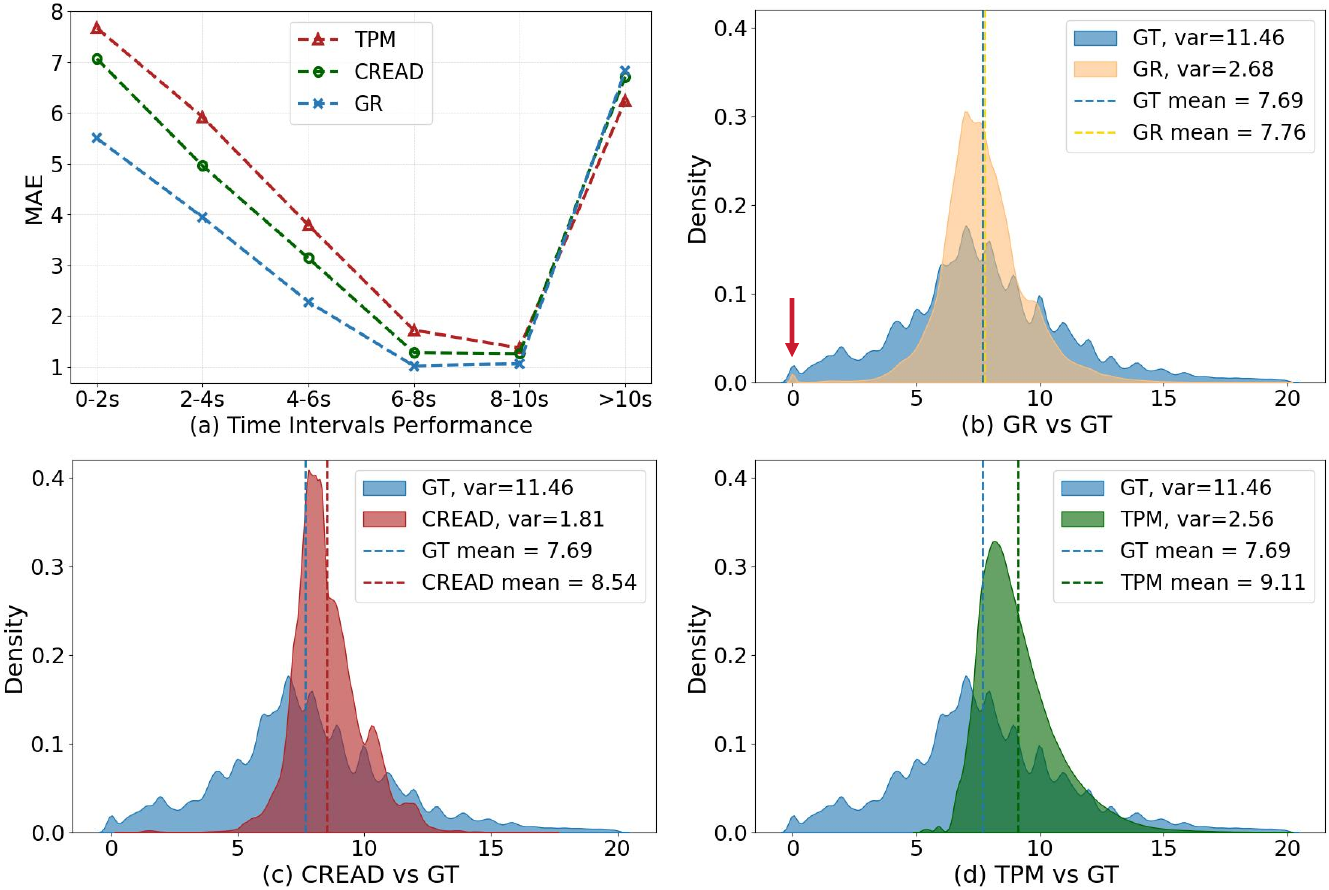}
    \caption{(a) Comparison of MAE on the KuaiRec dataset across videos with different watch time intervals. (b-d) The distribution comparison of predicted watch times among TPM, CREAD, and GR, compared to the Ground Truth (GT).}
    \label{fig:analysis}
\end{figure}

We analyze model performance across ground truth (GT) watch time intervals on KuaiRec, where approximately 80\% of videos have GT $\leq$10s. 
By splitting the range of watch time into 2-second long segments, Fig.~\ref{fig:analysis}(a) shows that GR significantly outperforms CREAD and TPM for videos with short and medium watch times, with slightly lower performance only in the last >10s interval, where it lags behind TPM.
For a more intuitive analysis, Fig.~\ref{fig:analysis}(b-d) visualizes the distributions of Ground Truth (GT) watch times and the predictions generated by different methods, alongside their means and variances.
Notably, the mean predicted watch time of GR closely aligns with the GT mean, whereas those of CREAD and TPM deviate significantly.
Regarding variance, GR exhibits the largest spread, while CREAD shows the smallest. This corresponds visually to CREAD's highly peaked distribution versus GR's broader and flatter curve, suggesting GR's capability to generate a more diverse and personalized set of predictions. 
Furthermore, GR is the only method that accurately predicts when GT is close to 0s, highlighting its flexibility afforded by its ability to output the <EOS> token in the first step.
Fig.~\ref{fig:analysis}(c) and (d) also visually confirm our hypothesis that CREAD and TPM tend to overestimate watch times, stemming from their rigid discretization structure where excessively large span values in tail buckets disproportionately amplify prediction errors, especially for videos with shorter watch times.
As shown in Fig.~\ref{fig:analysis}(d), the prediction distribution of TPM exhibits a notable skew towards higher values, attributable to the model's tendency (observed during case analysis) to learn probabilities greater than 0.5 at the root node of its tree structure.
This can result in an overall overestimation of the predicted outcomes, thereby explaining why GR's performance is marginally surpassed by TPM in the >10s interval. 
However, given the characteristic long-tail distribution of real-world watch time data, the superior overall performance and distributional fidelity achieved by GR represent a favorable trade-off for this minor discrepancy in the high-value range.


\begin{figure}[t]
    \centering
    \includegraphics[scale=0.32]{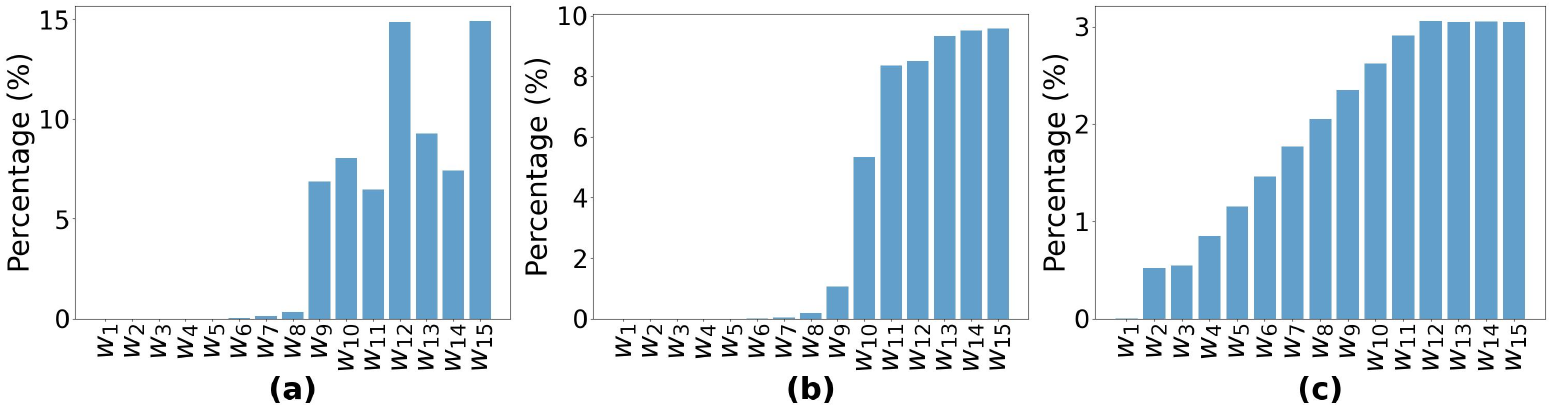}
    \caption{Token distribution comparison among vocabulary construction methods: (a) Manual, (b) Binary, (c) Dynamic.}
    \label{fig:token_frequency}
    \vspace{-1.5em}
\end{figure}

\subsection{Vocabulary Construction Analysis (RQ3)}
\label{sec:RQ3}
Here we examine the effect of the vocabulary construction method. 
Besides the proposed \textbf{Dynamic Quantile} algorithm, two commonly used methods are considered: \textbf{Manual} that designs the vocabulary based on experience, \textit{e.g.}, using values like 1ms, 3ms and 5ms, then scaling them by 10, 100, and so on until exceeding the maximum watch time in the dataset. 
\textbf{Binary} starts with the smallest unit of watching duration, i.e., milliseconds, as the first token, with each subsequent token being twice the value of its predecessor until exceeding the maximum watch time in the dataset.
Tab.~\ref{tab:vocabulary} presents the experimental results. We can see that the proposed dynamic quantile method outperforms the other two strategies. Notably, our method is nearly automatic, which makes it more efficient than the manual and binary vocabulary construction methods.

We further analyze token frequency distribution, i.e., counting the occurrences of each token in the vocabulary, and results are shown in Fig.~\ref{fig:token_frequency}. 
We sort all tokens in descending order according to frequencies and select the top 15 for analysis and comparison. 
We can see that in the binary method, nearly half of the tokens are scarcely used, while the manual method exhibits a highly imbalanced distribution. 
In contrast, our dynamic quantile method achieves a more balanced distribution, further validating the efficacy of the proposed algorithm.

\begin{table}[t]
\centering
\small
\caption{Ablation study on the strategy of curriculum learning (CL) with embedding mixup (EM).}
\label{tab:techear_force}
\resizebox{0.45\textwidth}{!}{
\begin{tabular}{c|c|cc|cc}
\hline
& \multirow{2}{*}{Method} & \multicolumn{2}{c|}{KuaiRec} & \multicolumn{2}{c}{CIKM16} \\
& & MAE~$\downarrow$ & XAUC~$\uparrow$ & MAE~$\downarrow$ & XAUC~$\uparrow$ \\
\hline
(a) & GR & \textbf{3.196} & \textbf{0.614} & \textbf{0.815} & \textbf{0.691} \\
\hline
(b) & w/o CLEM & 3.416 & 0.584 & 0.858 & 0.674 \\
(c) & EM with TF & 3.241 & 0.604 & 0.844 & 0.684  \\
(d) & CL w/o EM &  3.359 & 0.588 & 0.849 & 0.679 \\
\hline
(e) & {linear} & 3.205 & 0.613 & 0.818 & 0.690 \\
(f) & {exponential} & 3.211 & 0.613 & 0.819 & 0.690  \\
\hline
(g) & ${p=0.5}$ & 3.208 & 0.612 & 0.820 & 0.690  \\
(h) & ${p=0}$ & 3.283 & 0.593 & 0.846 & 0.681  \\
\hline
\end{tabular}
}
\end{table}

\subsection{Ablation study on Curriculum Learning with Embedding Mixup (RQ4)}
\label{sec:ablation}
To systematically evaluate the proposed Curriculum Learning with Embedding Mixup (CLEM) framework, we conduct controlled ablation experiments across three dimensions:(1) component effectiveness, (2) scheduling sensitivity, and (3) nonlinear decay impact. The experimental variants a re designed as follows:
\begin{itemize}[itemindent=0pt, left=4pt]
    \item \textbf{Component Analysis:}
    \begin{itemize}[left=8pt]
        \item {w/o CLEM}: Vanilla training using direct feature projection $\bm{E}[\hat{s}_i^{t-1},:] \rightarrow \hat{s}_i^t$ without curriculum scheduling or mixup.
        \item {EM with TF}: Embedding mixup with full teacher forcing (fixed sampling rate $p=1$).
        \item {CL w/o EM}: Curriculum learning without embedding mixup regularization.
    \end{itemize}
    \item \textbf{Decay Strategy Comparison:}
    \begin{itemize}[left=8pt]
        \item {Linear}: Linear sampling rate decay $p_t = 1 - \tau t$.
        \item {Exponential}: Exponential decay $p_t = e^{-\tau t}$.
    \end{itemize}
    \item \textbf{Sampling Rate Impact:}
    \begin{itemize}[left=8pt]
        \item {Fixed-0.5}: Constant sampling rate $p=0.5$.
        \item {Fixed-0}: Pure free-running mode ($p=0$).
    \end{itemize}
\end{itemize}

As shown in Tab.~\ref{tab:techear_force}, the full CLEM framework (Row a) demonstrates significant improvements over baseline configurations.
 Compared to the non-curriculum variant (Row c), curriculum learning alone provides a 1.656\% XAUC boost and 1.38\% MAE reduction on the KuaiRec dataset.
 Embedding mixup contributes more substantially: disabling mixup (Row d) degrades XAUC by 4.235\% and increases MAE by 4.853\%, highlighting its critical regularization role.
The sampling rate decay coefficients significantly impact both metrics. 
The proposed curriculum strategy achieves a gain of {2.65\%} in MAE and {3.42\%} in XAUC on KuaiRec, comparing row (a) with row (h). 
Although different nonlinear decay strategies yield similar results in terms of XAUC, they still improve MAE.
These findings indicate that the CLEM strategy improves the model's accuracy of watch time prediction.

\subsection{Performance on LTV Prediction Task (RQ5)}
\label{sec:ltv}
GR is a generalized regression framework. To rigorously evaluate its cross-task generalization capability, 
we conduct extended experiments on the Lifetime Value (LTV) prediction task under identical experimental protocols as
~\cite{weng2024optdist}. 
The evaluation employs two datasets:
Criteo-SSC\footnote{https://ailab.criteo.com/criteo-sponsored-search-conversion-log-dataset/} and Kaggle\footnote{https://www.kaggle.com/c/acquire-valued-shoppers-challenge}, with MAE and Spearman's rank correlation~(Spearman’s $\rho$) serving as performance metrics.
All baseline implementations strictly adhere to the configurations documented in~\cite{weng2024optdist}.

As shown in Tab.~\ref{tab:ltv}, GR achieves state-of-the-art performance with  \textit{relative improvements} of 17.66\% in MAE and 20.79\% in Spearman’s $\rho$ on Criteo-SSC over the previous best method OptDist~\cite{weng2024optdist}. Notably, these baselines include task-specific architectures with dedicated LTV prediction modules. 
The consistent superiority of GR across both point estimation (MAE) and ranking correlation ($\rho$) metrics provides empirical evidence for its inherent robustness and domain-agnostic characteristics. 


\begin{table}[t]
\centering
\caption{Performance comparison on LTV datasets.}
\label{tab:ltv}
\resizebox{0.48\textwidth}{!}{
\begin{tabular}{c|cc|cc}
\hline
\multirow{2}{*}{Method} & \multicolumn{2}{c|}{Criteo-SSC} & \multicolumn{2}{c}{Kaggle} \\
& MAE~$\downarrow$ & Spearman’s $\rho~\uparrow$ & MAE$~\downarrow$ &  Spearman’s $\rho~\uparrow$ \\
\hline
Two-stage~\cite{drachen2018or} & 21.719 & 0.2386 & 74.782 & 0.4313 \\
MTL-MSE~\cite{ma2018entire} & 21.190 & 0.2478 & 74.065 & 0.4329 \\
ZILN~\cite{wang2019deep} & 20.880 & 0.2434 & 72.528 & 0.5239 \\
MDME~\cite{li2022billion} & 16.598 & 0.2269 & 72.900 & 0.5163  \\
MDAN~\cite{liu2024mdan} & 20.030 & 0.2470 & 73.940 & 0.4367 \\
OptDist~\cite{weng2024optdist} & \underline{15.784} & \underline{0.2505} & \underline{70.929} & \underline{0.5249}  \\
\textbf{GR~(ours)} & \textbf{12.996} & \textbf{0.3026} & \textbf{67.035}  & \textbf{0.5334}  \\
\hline
\end{tabular}
}
\end{table}

\section{Conclusion}
This paper proposes a novel regression paradigm Generative Regression (\textbf{GR}) to accurately predict watch time, which addresses two key issues associated with existing ordinal regression (OR) methods.
First, 
OR struggles to accurately recover watch times due to discretization, with performance heavily reliant on the chosen time-binning strategy. 
Second, while OR implicitly constrains the probability distribution along the estimation path to exhibit a decreasing trend, existing methods have not fully leveraged this property.
GR builds upon autoregressive modeling and offers a promising exploration space. We also introduce embedding mixups and curriculum learning during training to accelerate model convergence.
Extensive online and offline experiments show that GR significantly outperforms the SOTA models. 
Additionally, 
our GR also surpasses the SOTA models in lifetime value (LTV) prediction, highlighting its potential as an effective general regression solution.

\bibliographystyle{ACM-Reference-Format}
\bibliography{main}

\end{document}